\pdfoutput=1

\documentclass[11pt]{article}

\usepackage{ACL2023}

\usepackage{times}
\usepackage{latexsym}

\usepackage[T1]{fontenc}

\usepackage{array} 
\usepackage{arydshln} 
\usepackage[utf8]{inputenc}

\usepackage{microtype}

\usepackage{inconsolata}

\usepackage{graphicx}
\usepackage{multirow}
\usepackage{booktabs}
\usepackage{paralist}
\usepackage{CJKutf8}

\usepackage{array,tabularx}
\usepackage{color} 
\definecolor{category}{RGB}{112,48,160}
\definecolor{aspect}{RGB}{68,114,196}
\definecolor{opinion}{RGB}{159,34,0}
\definecolor{sentiment}{RGB}{255,34,0}
\newcommand{\blue}[1]{{\color{blue}{#1}}}

\usepackage{amsmath,amsfonts,bm}
\usepackage{mathtools}

\def\eqref#1{equation~\ref{#1}}

\def\1{\bm{1}}

\def\rva{{\mathbf{a}}}
\def\rvb{{\mathbf{b}}}

\def\rvh{{\mathbf{h}}}

\def\rvo{{\mathbf{o}}}

\def\rvx{{\mathbf{x}}}

\def\rmC{{\mathbf{C}}}

\def\rmH{{\mathbf{H}}}

\def\rmP{{\mathbf{P}}}

\def\rmW{{\mathbf{W}}}

\def\rmY{{\mathbf{Y}}}

\DeclareMathAlphabet{\mathsfit}{\encodingdefault}{\sfdefault}{m}{sl}
\SetMathAlphabet{\mathsfit}{bold}{\encodingdefault}{\sfdefault}{bx}{n}

\def\gC{{\mathcal{C}}}

\def\gL{{\mathcal{L}}}

\def\gS{{\mathcal{S}}}

\def\sR{{\mathbb{R}}}

\def \NULL {\mbox{NULL}}
\title{{A Unified One-Step Solution for Aspect Sentiment Quad Prediction}}

\author{Junxian Zhou\textsuperscript{1}, Haiqin Yang\textsuperscript{2}\thanks{~~The corresponding author.} , Yuxuan He\textsuperscript{1}, Hao Mou\textsuperscript{1}, Junbo Yang\textsuperscript{1} \\
        \textsuperscript{1}DataStory, Guangzhou, China \\ 
        \textsuperscript{2}International Digital Economy Academy, Shenzhen, China \\ 
        \texttt{\{junius, heyuxuan, mouhao, junbo\}@datastory.com.cn}, \\             \texttt{hqyang@ieee.org}
        }

\begin{document}
\maketitle
\begin{abstract}

Aspect sentiment quad prediction (ASQP) is a challenging yet significant subtask in aspect-based sentiment analysis as it provides a complete aspect-level sentiment structure.  However, existing ASQP datasets are usually small and low-density, hindering technical advancement.  To expand the capacity, in this paper, we release two new datasets for ASQP, which contain the following characteristics: larger size, more words per sample, and higher density.  With such datasets, we unveil the shortcomings of existing strong ASQP baselines and therefore propose a unified one-step solution for ASQP, namely One-ASQP, to detect the aspect categories and to identify the aspect-opinion-sentiment (AOS) triplets simultaneously.  Our One-ASQP holds several unique advantages: (1) by separating ASQP into two subtasks and solving them independently and simultaneously, we can avoid error propagation in pipeline-based methods and overcome slow training and inference in generation-based methods; (2) by introducing sentiment-specific horns tagging schema in a token-pair-based two-dimensional matrix, we can exploit deeper interactions between sentiment elements and efficiently decode the AOS triplets; (3) we design ``[NULL]'' token can help us effectively identify the implicit aspects or opinions.  Experiments on two benchmark datasets and our released two datasets demonstrate the advantages of our One-ASQP. The two new datasets are publicly released at \url{https://www.github.com/Datastory-CN/ASQP-Datasets}.

\end{abstract}

\section{Introduction}\label{sec:intro}

\begin{table}[htp]
\footnotesize
\centering
\begin{tabular}{@{~}p{0.8cm}@{~~~}c@{~~~~}p{4.5cm}}
\hline
Task & Output & Example Output \\ \hline
ATE
& \{$a$\} & \{touch screen\} \\ \hline
ACD
& \{$c$\} & \{Screen\#Sensitivity\} \\ \hline
AOPE
& \{$(a, o)$\} & \{(touch screen, not sensitive)\} \\ \hline
ACSA
& \{$(c, s)$\} & \{(Screen\#Sensitivity, NEG)\} \\ \hline
{E2E-ABSA}
& \{$(a, s)$\} & \{(touch~screen, NEG)\} \\ \hline
ASTE
& \{$(a, o, s)$\} &  \{(touch~screen,  not sensitive, NEG)\} \\ \hline
TASD
& \{$(c, a, s)$\} & \{(Screen\#Sensitivity, touch~screen, NEG)\} \\ \hline
ASQP/ ACOS
& \multirow{2}{*}{\{$(c, a, o, s)$\}} & \{(Screen\#Sensitivity, touch screen, not sensitive, NEG)\} \\ \hline
\end{tabular}
\caption{The outputs of an example, ``touch screen is not sensitive'', for various ABSA tasks.  $a$, $c$, $o$, $s$, and NEG are defined in the first paragraph of Sec.~\ref{sec:intro}.
}
\label{tab:task}
\end{table}


Aspect-based sentiment analysis (ABSA) is a critical fine-grained opinion mining or sentiment analysis problem that aims to analyze and understand people's opinions or sentiments at the aspect level~\citep{DBLP:series/synthesis/2012Liu,DBLP:conf/semeval/PontikiGPPAM14,DBLP:journals/corr/abs-2203-01054}.  Typically, there are four fundamental sentiment elements in ABSA: (1) {\em aspect category} ($c$) defines the type of the concerned aspect; (2) {\em aspect term} ($a$) denotes the opinion target which is explicitly or implicitly mentioned in the given text; (3) {\em opinion term} ($o$) describes the sentiment towards the aspect; and (4) {\em sentiment polarity} ($s$) depicts the sentiment orientation.  For example, given an opinionated sentence, ``touch screen is not sensitive,'' we can obtain its $(c, a, o, s)$-quadruple as (``Screen\#Sensitivity'', ``touch screen'', ``not sensitive'', NEG), where NEG indicates the negative sentiment polarity.


Due to the rich usage of applications, numerous research efforts have been made on ABSA to predict or extract fine-grained sentiment elements~\citep{DBLP:conf/naacl/JiaoYKL19,DBLP:conf/semeval/PontikiGPPAM14,DBLP:conf/semeval/PontikiGPMA15,DBLP:conf/semeval/PontikiGPAMAAZQ16,DBLP:journals/corr/abs-2203-01054,DBLP:conf/ijcai/YangYDSZZ21}.  Based on the number of sentimental elements to be extracted, existing studies can be categorized into the following tasks: (1) {\em single term extraction} includes aspect term extraction (ATE)~\citep{DBLP:conf/emnlp/LiL17, DBLP:conf/acl/HeLND17}, aspect category detection (ACD)~\citep{DBLP:conf/acl/HeLND17,DBLP:conf/emnlp/0030TCL021};  
(2) {\em pair extraction} includes aspect-opinion pairwise extraction (AOPE)~\citep{DBLP:journals/taslp/YuJX19,DBLP:conf/emnlp/WuWP20}, aspect-category sentiment analysis (ACSA)~\citep{DBLP:conf/coling/CaiTZYX20,DBLP:conf/emnlp/DaiPCD20}, and End-to-End ABSA (E2E-ABSA)~\citep{DBLP:conf/emnlp/LiLWBZY19,DBLP:conf/acl/HeLND19} to extract the aspect and its sentiment; (3) {\em triplet extraction} includes aspect-sentiment triplet extraction (ASTE)~\citep{DBLP:conf/emnlp/MukherjeeNBB021,DBLP:conf/aaai/ChenWLW21}, and Target Aspect Sentiment Detection (TASD)~\citep{DBLP:conf/aaai/WanYDLQP20}; (4) {\em quadruple extraction} includes aspect-category-opinion-sentiment (ACOS) quadruple extraction~\citep{DBLP:conf/acl/CaiXY20} and aspect sentiment quad prediction (ASQP)~\citep{DBLP:conf/emnlp/ZhangD0YBL21}.  ACOS and ASQP are the same tasks, which aim to extract all aspect-category-opinion-sentiment quadruples per sample.  Since ASQP covers the whole task name, we use ASQP to denote the ABSA quadruple extraction task.  Table~\ref{tab:task} summarizes an example of the outputs of various ABSA tasks.  

This paper focuses on ASQP because it provides a complete aspect-level sentiment analysis~\citep{DBLP:journals/corr/abs-2203-01054}.  We first observe that existing ASQP datasets are crawled from only one source and are small with low-density~\citep{DBLP:conf/acl/CaiXY20,DBLP:conf/emnlp/ZhangD0YBL21}.  For example, the maximum sample size is around 4,000, while the maximum number of quadruples per sample is around 1.6.  This limits the technical development of ASQP.  Second, ASQP includes two extraction subtasks  (aspect extraction and opinion extraction) and two classification subtasks (category classification and sentiment classification).  Modeling the four subtasks simultaneously is challenging, especially when the quadruples contain implicit aspects or opinions~\citep{DBLP:conf/acl/CaiXY20}.  Though existing studies can resolve ASQP via pipeline-based~\citep{DBLP:conf/acl/CaiXY20} or generation-based methods~\citep{DBLP:conf/emnlp/ZhangD0YBL21,DBLP:conf/acl/MaoSYZC22,DBLP:conf/ijcai/BaoWJXL22,DBLP:conf/coling/GaoFLLLLBY22}, they suffer from different shortcomings, i.e., pipeline-based methods tend to yield error propagation while generation-based methods perform slowly in training and inference. 

To tackle the above challenges, we first construct two datasets, \textbf{en-Phone} and \textbf{zh-FoodBeverage}, to expand the capacity of datasets.  {en-Phone} is an English ASQP dataset in the cell phone domain collected from several e-commercial platforms, while {zh-FoodBeverage} is the first Chinese ASQP dataset collected from multiple sources under the categories of Food and Beverage.  Compared to the existing ASQP datasets, our datasets have 1.75 to 4.19 times more samples and a higher quadruple density of 1.3 to 1.8.  This achievement is a result of our meticulous definition and adherence to annotation guidelines, which allow us to obtain more fine-grained quadruples.


After investigating strong ASQP baselines, we observed a decline in performance on our newly released dataset. This finding, coupled with the shortcomings of the existing baselines, motivated us to develop a novel one-step solution for ASQP, namely One-ASQP.  As illustrated in Fig.~\ref{fig:model}, our One-ASQP adopts a shared encoder from a pre-trained language model (LM) and resolves two tasks, aspect category detection (ACD) and aspect-opinion-sentiment co-extraction (AOSC)  simultaneously.  ACD is implemented by a multi-class classifier and AOSC is fulfilled by a token-pair-based two-dimensional (2D) matrix with the sentiment-specific horns tagging schema, a popular technique borrowed from the joint entity and relation extraction~\citep{DBLP:conf/coling/WangYZLZS20,DBLP:conf/aaai/ShangHM22}.  The two tasks are trained independently and simultaneously, allowing us to avoid error propagation and overcome slow training and inferring in generation-based methods.  Moreover, we also design a unique token, ``[NULL]'', appending at the beginning of the input, which can help us to identify implicit aspects or opinions effectively.    

\begin{table*}[htp]
\scriptsize
\centering
\begin{tabular}{
l@{~~~~}r@{~~~~}r@{~~~~}r@{~~~~}r@{~~~~}r@{~~~~}r@{~~~~}r@{~~~~}c@{~~~~}c@{~~~~}c@{~~~~}c@{~~~~}c@{~~~~}c@{~~~~}c
}
\hline
 & \multicolumn{1}{c}{\#s} & \#w/s
 & \#$c$ & \multicolumn{1}{c}{\#q} & \#q/s
 & EA\&EO & EA\&IO & IA\&EO & IA\&IO & {\#NEG} & {\#NEU} & {\#POS} & Avg. \#w/$a$ & Avg. \#w/$o$ \\ \hline
Restaurant-ACOS & 2,286 & 15.11 & 13 & 3,661 & 1.60 & 2,431 & 350 & 530 & 350 & 1,007 & 151 & 2,503 & 1.46 & 1.20 \\ \hline
Laptop-ACOS & 4,076 & 15.73 & 121 & 5,773 & 1.42 & 3,278 & 1,241 & 912 & 342 & 1,879 & 316 & 3,578 & 1.40 & 1.09 \\ 
\hline\hline
en-Phone & 7,115 & 25.78 & 88 & 15,884 & 2.23 & 13,160 & 2,724 & - & - & 3,751 & 571 & 11,562 & 1.73 & 1.98 \\ \hline
zh-FoodBeverage & 9,570 & 193.95 & 138 & 24,973 & 2.61 & 17,407 & 7,566 & - & - & 6,778 & - & 18,195 & 2.60 & 2.04 
 \\ \hline
\end{tabular}
\caption{Data statistics for the ASQP task.  \# denotes the number of corresponding elements.  s, w, $c$, q stand for samples, words, categories, and quadruples, respectively.  EA, EO, IA, and IO denote explicit aspect, explicit opinion, implicit aspect, and implicit opinion, respectively. ``-'' means this item is not included.}
\label{tab:datasets}
\end{table*}

Our contributions are three-fold: (1) We construct two new ASQP datasets consisting of more fine-grained samples with higher quadruple density while covering more domains and languages.  Significantly, the released zh-FoodBeverage dataset is the first Chinese ASQP dataset, which provides opportunities to investigate potential technologies in a multi-lingual context for ASQP.  (2) We propose One-ASQP to simultaneously detect aspect categories and co-extract aspect-opinion-sentiment triplets.  One-ASQP can absorb deeper interactions between sentiment elements without error propagation and conquer slow performance in generation-based methods.  Moreover, the delicately designed ``[NULL]'' token helps us to identify implicit aspects or opinions effectively.  (3) We conducted extensive experiments demonstrating that One-ASQP is efficient and outperforms the state-of-the-art baselines in certain scenarios.

\section{Datasets}
We construct two datasets~\footnote{{More details are provided in Appendix~\ref{app:datasets}.}} to expand the capacity of existing ASQP datasets.

\subsection{Source} 

{\bf en-Phone} is an English dataset collected from reviews on multiple e-commerce platforms in July and August of 2021, covering 12 cell phone brands.  To increase the complexity and the quadruple density of the dataset, we deliver the following filtering steps: (1) applying the LangID toolkit~\footnote{\url{https://pypi.org/project/langid/}} to filter out comments whose body content is not in English; (2) filtering out samples with less than $8$ valid tokens.  {\bf zh-FoodBeverage} is the first Chinese ASQP dataset, collected from Chinese comments in multiple sources in the years 2019-2021 under the categories of Food and Beverage.  We clean the data by (1) filtering out samples with lengths less than $8$ and greater than $4000$; (2) filtering out the samples with valid Chinese characters less than 70\%; (3) {filtering out ad texts by a classifier which is trained by marketing texts with 90\% classification accuracy. 
}


\subsection{Annotation}

A team of professional labelers is asked to label the texts following the guidelines in Appendix~\ref{app:guideline}.  {Two annotators individually annotate the same sample by our internal labeled system.}  The strict quadruple matching F1 score between two annotators is 77.23\%, which implies a substantial agreement between two annotators~\citep{DBLP:conf/coling/KimK18}.  In case of disagreement, the project leader will be asked to make the final decision.  Some typical examples are shown in Table~\ref{tab:ex_illustration}.  

\begin{figure*}[ht] 
\centering 
\includegraphics[scale=0.135]{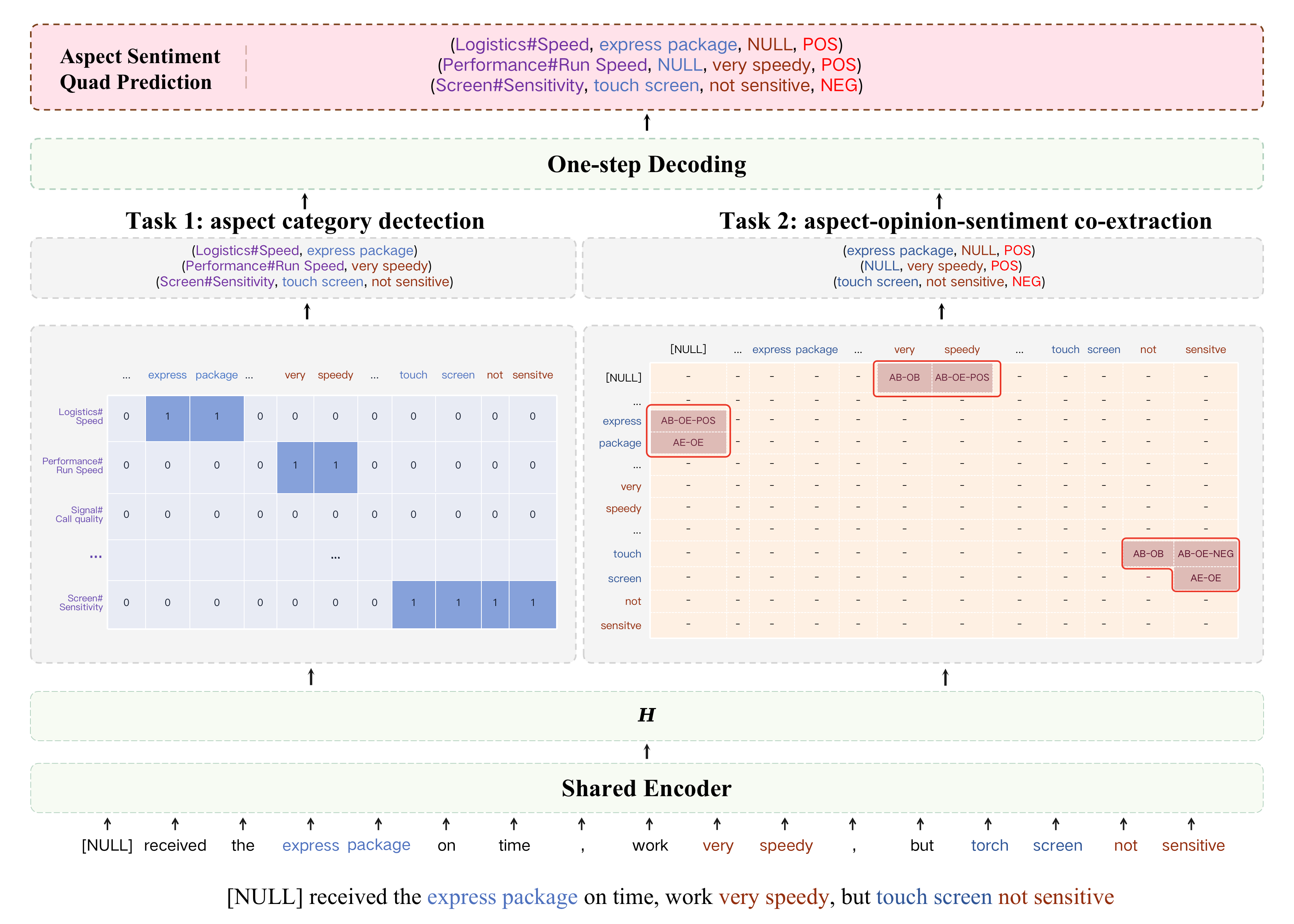}
\caption{The structure of our One-ASQP: solving ACD and AOSC simultaneously.  ACD is implemented by a multi-class classifier.  AOSC is fulfilled by a token-pair-based 2D matrix with sentiment-specific horns tagging.  The results in the row of ``[NULL]'' indicate no aspect for the opinion of ``very speedy''. In contrast, the results in the column of ``[NULL]'' imply no opinion for the aspect of ``express package''.
} 
\label{fig:model} 
\end{figure*}

\subsection{Statistics and Analysis}
{
Table~\ref{tab:datasets} reports the statistics of two existing ASQP benchmark datasets and our released datasets for comparison.  \textbf{en-Phone} contains 7,115 samples with 15,884 quadruples while \textbf{zh-FoodBeverage} contains 9,570 samples with 24,973 quadruples.  The size and the number of quadruples are significantly larger than the current largest ASQP benchmark dataset, i.e., Laptop-ACOS.  The statistics show that our released datasets contain unique characteristics and are denser than existing Restaurant-ACOS and Laptop-ACOS: (1) the number of words per sample is 25.78 and 193.95 for en-Phone and zh-FoodBeverage, respectively, while the number of quadruples per sample is 2.23 and 2.61 for en-Phone and zh-FoodBeverage accordingly.  This shows that en-Phone and zh-FoodBeverage are much denser than existing ASQP datasets; (2) based on the annotation guidelines, we only label opinionated sentences with explicit aspects.  Moreover, due to commercial considerations, we exclude sentences with neutral sentiment in zh-FoodBeverage; (3) here, we define more fine-grained aspects and opinions than existing ASQP datasets; see more examples in Appendix~\ref{app:datasets}.  Consequently, we attain a longer average length per aspect and per opinion, as reported in the last two columns of Table~\ref{tab:datasets}.

\section{Methodology}
\subsection{ASQP Formulation}
Given an opinionated sentence $\rvx$, ASQP is to predict all aspect-level sentiment quadruples $\{(c, a, o, s)\}$, which corresponds to the aspect category, aspect term, opinion term, and sentiment polarity, respectively.  The aspect category $c$ belongs to a category set $\gC$; the aspect term $a$ and the opinion term $o$ are typically text spans in $\rvx$ while they can be null if the target is not explicitly mentioned, i.e., $a\in V_{\rvx}\cup\{\emptyset\}$ and $o\in V_{\rvx}\cup\{\emptyset\}$, where $V_{\rvx}$ denotes the set containing all possible continuous spans of $\rvx$.  The sentiment polarity $s$ belongs to one of the sentiment classes, SENTIMENT=\{POS, NEU, NEG\}, which corresponds to the positive, neutral, and negative sentiment, respectively. 


\subsection{One-ASQP}\label{sec:model}

Our One-ASQP resolves two subtasks, ACD and AOSC, simultaneously, where ACD seeks a classifier to determine the aspect categories, and AOSC is to extract all $(a, o, s)$-triplets.  

Given $\rvx$ with $n$-tokens, we construct the input as follows:
\begin{equation}\label{eq:input}
 [\NULL]\,x_1\, x_2\, \ldots\, x_n,
\end{equation}
where the token $[\NULL]$ is introduced to detect implicit aspects or opinions; see more details in Sec.~\ref{sec:AOSC}.  Now, via a pre-trained LM, both tasks share a common encoder to get the representations:
\begin{equation}\label{eq:input_representation}
\rmH = \rvh_{\tiny\NULL}\, \rvh_1\, \rvh_2\, \ldots\, \rvh_n \in \sR^{d\times(n+1)},
\end{equation}
where $d$ is the token representation size.  

\subsubsection{Aspect Category Detection}\label{sec:ACD}
We apply a classifier to predict the probability of category detection: 
\begin{equation}\label{eq:ACD}
\rmC = \mbox{sigmoid}(\rmW_2(\mbox{RELU}(\rmW_1 \rmH + \rvb_1 ))),
\end{equation}
where $\rmW_1 \in \sR^{d\times d}$, $\rvb_1 \in \sR^{d}$ , $\rmW_2 \in \sR^{|\gC|\times d}$.  Here, $|\gC|$ is the number of categories in $\gC$.  Hence, $\rmC\in \sR^{|\gC|\times (n+1)}$, where $C_{ij}$ indicates the probability of the $i$-th token to the $j$-th category.  


\subsubsection{AOSC}\label{sec:AOSC}

We tackle AOSC via a token-pair-based 2D matrix with the sentiment-specific horns tagging schema to determine the positions of aspect-opinion pairs and their sentiment polarity.  
{
\paragraph{Tagging} We define four types of tags: (1) \textbf{AB-OB} denotes the cell for the beginning position of an aspect-opinion pair.  For example, as (``touch screen'',  ``not sensitive'') is an aspect-opinion pair, the cell corresponding to (``touch'', ``not'') in the 2D matrix is marked by ``AB-OB''. (2) \textbf{AE-OE} indicates the cell for the end position of an aspect-opinion pair.  Hence, the cell of (``screen'', ``sensitive'') is marked by ``AE-OE''.  (3) \textbf{AB-OE-*SENTIMENT} defines a cell with its sentiment polarity, where the row position denotes the beginning of an aspect and the column position denotes the end of an opinion.  Hence,  the cell of (``touch'', ``sensitive'') is tagged by ``AB-OE-NEG''.   As SENTIMENT consists of three types of sentiment polarity, there are three cases in AB-OE-*SENTIMENT.  
(4) {\bf ``-''} denotes the cell other than the above three types.  Hence, we have five types of unique tags, \{AB-OB, AE-OE, AB-OE-POS, AB-OE-NEU, AB-OE-NEG\}. 

\paragraph{Triplet Decoding} Since the tagged 2D matrix has marked the boundary tokens of all aspect-opinion pairs and their sentiment polarity, we can decode the triples easily.  First, by scanning the 2D matrix column-by-column, we can determine the text spans of an aspect, starting with ``AB-OE-*SENTIMENT'' and ending with ``AE-OE''.  Similarly, by scanning the 2D matrix row-by-row, we can get the text spans of an opinion, which start from ``AB-OB'' and end with ``AB-OE-*SENTIMENT''.  Finally, the sentiment polarity can be easily determined by ``AB-OE-*SENTIMENT''. 

\paragraph{Implicit Aspects/Opinions Extraction}
Detecting implicit aspects or opinions is critical in ASQP~\citep{DBLP:conf/acl/CaiXY20}.  Here, we append the ``[NULL]'' token at the beginning of the input.  Our One-ASQP can then easily determine the cases of Implicit Aspects and Explicit Opinions (IA\&EO) and Explicit Aspects and Implicit Opinions (EA\&IO).  The whole procedure is similar to the above triplet decoding: when the text spans at the row of ``[NULL]'' start from ``AB-OB'' and end with ``AB-OE-*SENTIMENT'', we can obtain an explicit opinion without aspect.  Meanwhile, when the text spans at the column of ``[NULL]'' start from ``AB-OE-*SENTIMENT'' and ends with ``AE-OE'', we can obtain an explicit aspect without opinion.  As shown in Fig.~\ref{fig:model}, we can quickly obtain the corresponding aspect-opinion pairs as ``(NULL, very speedy)'' and ``(express package, NULL)''.  The sentiment polarity can also be determined by ``AB-OE-*SENTIMENT'' accordingly.  Although the current setting for IA\&IO cannot be solved directly, it is possible to resolve it in two steps. First, we can identify IA\&IO using tools such as Extract-Classify-ACOS~\citep{DBLP:conf/acl/CaiXY20}. Then, we can classify aspect categories and sentiment polarity. However, a unified solution with One-ASQP is left for future work.

\paragraph{Tagging Score}

Given $\rmH$, we compute the probabilities of the $(i, j)$-th cell to the corresponding tags by: 
\begin{align}
  \rva_{i}&=\rmW_{a}\rvh_i+\rvb_{a}, \\
  \rvo_{j}&=\rmW_{o}\rvh_j+\rvb_{o}, \\ \label{eq:tag_score}
  \rmP_{ij}&=\mbox{sigmoid}(\rva^T_{i}\rvo_{j}) \in \sR^{5}
\end{align}
where $\rmW_{a} \in \sR^{D\times d}$ and $\rmW_{o} \in \sR^{D\times d}$ are the weight matrices for the aspect token and the opinion token, respectively, $\rvb_{a} \in \sR^{D}$ and $\rvb_{o}  \in \sR^{D}$ are the biases for the aspect token and the opinion token, respectively.  $D$ is the hidden variable size set to 400 as default.  

\subsection{Training Procedure}

\paragraph{Training} We train ACD and AOSC jointly by minimizing the following loss function: 
\begin{equation}\label{eq:One-ASQP_loss}
\gL_{\scriptscriptstyle total} = \alpha\mathcal{L}_{\scriptscriptstyle ACD} + \beta\mathcal{L}_{\scriptscriptstyle AOSC},
\end{equation}
where $\alpha$ and $\beta$ are trade-off constants set to 1 for simplicity.  The ACD loss $\gL_{\scriptscriptstyle ACD}$ and the AOSC loss $\gL_{\scriptscriptstyle AOSC}$ are two cross-entropy losses defined as follows:  
\begin{align}
& \gL_{\scriptscriptstyle ACD} = -\frac{1}{n\times |\gC|} \times \\\nonumber
&\sum_{i=1}^{n} \sum_{j=0}^{|\gC|-1}\{y^{\gC}_{ij}\log C_{ij}+(1-y^{\gC}_{ij})\log(1-C_{ij})\},\\
& \gL_{\scriptscriptstyle AOSC} = -\frac{1}{(n+1)\times (n+1)\times 5} \times\\\nonumber
& \sum_{i=0}^{n} \sum_{j=0}^{n} \{\rmY_{ij}^t\log{\rmP_{ij}} + (1-\rmY_{ij}^t)\log(1-\rmP_{ij})\},  
\end{align}
where $C_{ij}$ is the predicted category computed by Eq.~(\ref{eq:ACD}), $y^{\gC}_{ij}\in\{0, 1\}$ and it is 1 when the $i$-th token is assigned to the $j$-th category and 0 otherwise.  $\rmP_{ij}$ is the predicted tagging score computed by Eq.~(\ref{eq:tag_score}) for all five types of tags while $\rmY_{ij}\in\sR^5$ is the ground-truth one-hot encoding. 

During training, we implement the negative sampling strategy as~\citep{DBLP:conf/iclr/LiL021} to improve the performance of our One-ASQP on unlabeled quadruples.  We set the negative sampling rate to 0.4, a suggested range in~\citep{DBLP:conf/iclr/LiL021} that has yielded good results.  Specifically, to minimize the loss in Eq.(\ref{eq:One-ASQP_loss}), we randomly sample 40\% of unlabeled entries as negative instances, which correspond to `0' in ACD and `-' in AOSC, as shown in Fig.\ref{fig:model}.

 }





\subsection{Quadruples Decoding}
After attaining the model, we can obtain the category sequences of ACD and the AOS triplets in the AOSC matrix simultaneously.  We then decode the quadruples in one step via their common terms.  For example, as shown in Fig.~\ref{fig:model}, we can merge (Logistics\#Speed, express package) and (express package, NULL, POS) via the common aspect term, ``express package'', and obtain the quadruple (Logistics\#Speed, express package, NULL, POS).  

Overall, our One-ASQP consists of two independent tasks, ACD and AOSC.  Their outputs only share in the final decoding stage and do not rely on each other during training as the pipeline-based methods need.  This allows us to train the model efficiently and decode the results consistently in both training and test.

\section{Experimental Setup}
\paragraph{Datasets} 
{
We conduct the experiments on four datasets in Table~\ref{tab:datasets}.  {For Restaurant-ACOS and Laptop-ACOS, we apply the original splitting on the training, validation, and test sets~\cite{DBLP:conf/acl/CaiXY20}.  For en-Phone and zh-FoodBeverage, the splitting ratio is 7:1.5:1.5 for training, validation, and test, respectively.

}

\paragraph{Evaluation Metrics} We employ F1 scores as the main evaluation metric and also report the corresponding \textbf{P}recision and \textbf{R}ecall scores.  A sentiment quad prediction is counted as correct if and only if all the predicted elements are exactly the same as the gold labels.  The time cost is also recorded to demonstrate the efficiency of One-ASQP.  

\paragraph{Implementation Details}  
{One-ASQP is implemented by PyTorch 1.13.1.  All experiments are run on a workstation with an Intel Xeon E5-2678 v3@2.50GHz CPU, 256G memory, a single A5000 GPU, and Ubuntu 20.04.  For English datasets, we adopt LMs of DeBERTaV3-base and DeBERTaV3-large~\citep{he2021debertav3}, which contain 12 layers with a hidden size of 768 and 24 layers with a hidden size of 1024, respectively.  {For the Chinese dataset, we adopt MacBERT~\citep{DBLP:conf/emnlp/CuiC000H20}, a Chinese LM with the same structure as DeBERTaV3.} 
For the English datasets, the maximum token length is set to 128 as the maximum average word length is only 25.78, as shown in Table~\ref{tab:datasets}.  For the zh-FoodBeverage, the maximum token length is 256.  The batch size and learning rate for all experiments are [32, 3e-5] as they can perform well.  We monitor the F1 score on the validation set and terminate the training when no score drops for four epochs.  Finally, we report the scores on the test set by the best model on the validation set.
} 
\begin{table}[htpb]
\scriptsize
\centering
\begin{tabular}{l@{~~}|c@{~~}c@{~~}c@{~~~}|c@{~~}c@{~~}c}
\hline
\multirow{2}{*}{\textbf{Method}}
& \multicolumn{3}{c@{~~}|@{~~}}{\textbf{Restaurant-ACOS}} & \multicolumn{3}{c}{\textbf{Laptop-ACOS}} \\ \cline{2-7} 
& {{P}} & {{R}} & {{F1}} & {{P}} & {{R}} & {{F1}} \\ \hline
DP-ACOS & 34.67 & 15.08 & 21.04 & 13.04 & 0.57 & 8.00 \\
JET-ACOS & 59.81 & 28.94 & 39.01 & 44.52 & 16.25 & 23.81 \\
TAS-BERT-ACOS & 26.29 & 46.29 & 33.53 & \textbf{47.15} & 19.22 & 27.31 \\
Extract-Classify-ACOS & 38.54 & 52.96 & 44.61 & 45.56 & 29.48 & 35.80 \\\hdashline
BARTABSA & 56.62 & 55.35 & 55.98 & 41.65 & 40.46 & 41.05 \\
GAS & 60.69 & 58.52 & 59.59 & 41.60 & 42.75 & 42.17 \\
Paraphrase & 58.98 & 59.11 & 59.04 & 41.77 & \textbf{45.04} & 43.34 \\
Seq2Path & - & - & 58.41 & - & - & 42.97 \\
GEN-SCL-NAT & - & - & 62.62 & - & - & 45.16 \\ 
OTG & 63.96 & \textbf{61.74} & \textbf{62.83} & 46.11 & 44.79 & \textbf{45.44} \\
\hline
One-ASQP (base) & 62.60 & 57.21 & 59.78 & 42.83 & 40.00 & 41.37 \\
One-ASQP (large) & \textbf{65.91} & 56.24 & 60.69 & 43.80 & 39.54 & 41.56 \\ \hline
\end{tabular}
\caption{Results of Restaurant-ACOS and Laptop-ACOS.  Scores are averaged over 5 runs with different seeds. }
\label{exp:main_result1}
\end{table}
\paragraph{Baselines}
We compare our One-ASQP with strong baselines: (1) {\em pipeline-based methods} consist of four methods, i.e., \textbf{DP-ACOS}, \textbf{JET-ACOS}, \textbf{TAS-BERT-ACOS}, and \textbf{Extract-Classify-ACOS}, which are all proposed in~\citep{DBLP:conf/acl/CaiXY20}; (2) {\em generation-based methods} include BART for ABSA (\textbf{BARTABSA})~\citep{DBLP:conf/acl/YanDJQ020}, Generative Aspect-based Sentiment analysis  (\textbf{GAS})~\citep{DBLP:conf/acl/Zhang0DBL20}, \textbf{Paraphrase} generation for ASQP~\citep{DBLP:conf/emnlp/ZhangD0YBL21}, \textbf{Seq2Path}~\citep{DBLP:conf/acl/MaoSYZC22}, \textbf{GEN-SCL-NAT}~\citep{DBLP:journals/corr/abs-2211-07743}, and ABSA with Opinion Tree Generation (\textbf{OTG})~\citep{DBLP:conf/ijcai/BaoWJXL22}.  

\section{Results and Discussions}
\subsection{Main Results}
\begin{table}[htp]
\scriptsize
\centering
\begin{tabular}{l@{~~}|c@{~~}c@{~~}c@{~~~}|c@{~~}c@{~~}c}
\hline

\multirow{2}{*}{\textbf{Method}}
& \multicolumn{3}{c@{~~}|@{~~}}{\textbf{en-Phone}} & \multicolumn{3}{c}{\textbf{zh-FoodBeverage}} \\ \cline{2-7} 
& {{P}} & {{R}} & {{F1}} & {{P}} & {{R}} & {{F1}} \\ \hline
Extract-Classify-ACOS & 31.28 & 33.23 & 32.23 & 41.39 & 32.53 & 36.43 \\
Paraphrase & 46.72 & 49.84 & 48.23 & 52.74 & 50.47 & 51.58 \\
GEN-SCL-NAT & 45.16 & \textbf{51.56} & 48.15 & 54.28 & 48.95 & 51.48 \\ \hline
One-ASQP (base) & \textbf{57.90} & 49.86 & 53.58 & 56.51 & \textbf{59.13} & 57.79 \\
One-ASQP (large) & 57.42 & 50.96 & \textbf{54.00} & \textbf{60.96} & 56.24 & \textbf{58.51} \\ \hline
\end{tabular}
\caption{Results of en-Phone and zh-FoodBeverage.  Scores are averaged over five runs with different seeds.}
\label{tab:Phone_FoodBeverage}
\end{table}

Table~\ref{exp:main_result1} reports the comparison results on two existing ASQP datasets.  Since all methods apply the same splitting on these two datasets, we copy the results of baselines from corresponding references.  The results show that: (1) Generation-based methods gain significant improvement over pipeline-based methods as pipeline-based methods tend to propagate the errors.  (2) Regarding generation-based methods, OTG attains the best performance on the F1 score.  The exceptional performance may come from integrating various features, e.g., syntax and semantic information, for forming the opinion tree structure~\citep{DBLP:conf/ijcai/BaoWJXL22}.  (3) Our One-ASQP is competitive with generation-based methods.  By checking the LM sizes, we know that the generation-based baselines except BARTABSA apply T5-base as the LM, which consists of 220M parameters.  In comparison, our One-ASQP model utilizes DeBERTaV3, which consists of only 86M and 304M backbone parameters for its base and large versions, respectively. The compact model parameter size is a crucial advantage of our approach.  However, on the Restaurant-ACOS and Laptop-ACOS datasets, One-ASQP falls slightly behind some generation-based methods that can take advantage of the semantics of sentiment elements by generating natural language labels.  In contrast, One-ASQP maps each label to a specific symbol, similar to the numerical indexing in classification models.  Unfortunately, the limited quantity of these datasets prevents our One-ASQP model from achieving optimal performance.


{We further conduct experiments on en-Phone and zh-FoodBeverage and compare our One-ASQP with three strong baselines, Extract-Classify-ACOS, Paraphrase, and GEN-SCL-NAT.  We select them because Extract-Classify-ACOS is the best pipeline-based method. Furthermore, Paraphrase and GEN-SCL-NAT are two strong generation-based baselines releasing source codes, which is easier for replication.}  Results in Table~\ref{tab:Phone_FoodBeverage} are averaged by five runs with different random seeds and show that our One-ASQP, even when adopting the base LM version, outperforms three strong baselines.  We conjecture that good performance comes from two reasons: (1) The newly-released datasets contain higher quadruple density and fine-grained sentiment quadruples.  This increases the task difficulty and amplifies the order issue in generation-based methods~\cite{DBLP:conf/acl/MaoSYZC22}, i.e., the orders between the generated quads do not naturally exist, or the generation of the current quads should not condition on the previous ones.  More evaluation tests are provided in Sec.~\ref{sec:effect_density}.} (2) The number of categories in the new datasets is much larger than Restaurant-ACOS and Laptop-ACOS.  This also increases the search space, which tends to yield generation bias, i.e., the generated tokens neither come from the original text nor the pre-defined categories and sentiments.  Overall, the results  demonstrate the significance of our released datasets for further technical development.

\begin{table}[htp]
\scriptsize
\centering
\begin{tabular}{l@{~}|c@{~}c@{~}|c@{~}c@{~}}
\hline
\multirow{2}{*}{\textbf{Method}} & \multicolumn{2}{c|}{\textbf{Restaurant-ACOS}} & \multicolumn{2}{c}{\textbf{en-Phone}} \\ \cline{2-5} 
& {{Train}} & {{Inference}} & {{Train}} & {{Inference}} \\ \hline
Extract-Classify-ACOS & 38.43 & 14.79 & 158.34 & 25.23 \\
Paraphrase & 30.52 & 58.23 & 99.23 & 160.56  \\
GEN-SCL-NAT & 35.32 & 61.64 & 104.23 & 175.23 \\ \hline
OneASQP (base) & 11.23 & 6.34 (29.35) & 32.23 & 6.32 (35.45) \\
OneASQP (large) &  17.63 & 14.63 (44.62) & 105.23 & 10.34 (61.23) \\
\hline
\end{tabular}
\caption{Time cost (seconds) on Restaurant-ACOS and en-Phone.  For a fair comparison with baselines, we record the inference time of our One-ASQP with the batch size of 1 and report them in the round bracket. }
\label{tab:time}
\end{table}

Table~\ref{tab:time} reports the time cost (in seconds) of training in one epoch and inferring the models on Restaurant-ACOS and en-Phone; more results are in Appendix~\ref{app:efficiency}.  The results show that our One-ASQP is much more efficient than the strong baselines as Extract-Classify-ACOS needs to encode twice and Paraphrase can only decode the token sequentially.  To provide a fair comparison, we set the batch size to 1 and show the inference time in the round bracket.  The overall results show that our One-ASQP is more efficient than the baselines.  Our One-ASQP can infer the quadruples parallel, which is much favorite for real-world deployment. 

\subsection{Effect of Handling Implicit Aspects/Opinions}
\begin{table*}[htpb]
\footnotesize
\centering
\begin{tabular}{l@{~~}|c@{~~}c@{~~}c@{~~~}|c@{~~}c@{~~}c@{~~~}|c@{~~}c@{~~}|c@{~~}c@{~~}}
\hline
{\multirow{2}{*}{\textbf{Method}}} & \multicolumn{3}{c|}{\textbf{Restaurant-ACOS}} & \multicolumn{3}{c|}{\textbf{Laptop-ACOS}} & \multicolumn{2}{c|}{\textbf{en-Phone}} & \multicolumn{2}{c}{\textbf{zh-FoodBeverage}} \\ \cline{2-11} 
 & {{EA\&EO}} & {{EA\&IO}} & {{IA\&EO}} & {{EA\&EO}} & {{EA\&IO}} & {{IA\&EO}} & {{EA\&EO}} & {{EA\&IO}} & {{EA\&EO}} & {{EA\&IO}} \\ \hline
Extract-Classify & 45.0 & 23.9 & 34.7 & 35.4 & 16.8 & 39.0 & 35.2 & 24.2 & 37.2 & 33.3 \\
Paraphrase & 65.4 & 45.6 & 53.3 & 45.7 & 33.0 & 51.0 & 49.1 & 45.6 & 50.9 & 49.9 \\
GEN-SCL-NAT & \textbf{66.5} & \textbf{46.2} & 56.5 & \textbf{45.8} & \textbf{34.3} & \textbf{54.0} & 50.1 & 45.4& 50.9 & 49.9 \\\hline
One-ASQP & 66.3 & 31.1 & \textbf{64.2} & 44.4 & 26.7 & 53.5 & \textbf{54.8} & \textbf{52.9} & \textbf{55.4} & \textbf{59.8} \\ \hline
\end{tabular}
\caption{Breakdown performance (F1 scores) to depict the ability to handle implicit aspects or opinions.  E and I stand for Explicit and Implicit, respectively, while A and O denote Aspect and Opinion, respectively.}
\label{tab:implicit}
\end{table*}
{Table~\ref{tab:implicit} reports the breakdown performance of the methods in addressing the implicit aspects/opinions problem.  The results show that {(1) the generation-based baseline, GEN-SCL-NAT, handles EA\&IO better than our One-ASQP when the quadruple density is low.  Accordingly, One-ASQP performs much better than GEN-SCL-NAT on IA\&EO in Restaurant-ACOS.  GEN-SCL-NA performs worse in IA\&EO may be because the generated decoding space of explicit opinions is huge compared to explicit aspects. (2) In en-Phone and zh-FoodBeverage, One-ASQP consistently outperforms all baselines on EA\&EO and EA\&IO.  Our One-ASQP is superior in handling implicit opinions when the datasets are more fine-grained. 
}}

\subsection{Ablation Study on ADC and AOSC}\label{sec:ablation}
{To demonstrate the beneficial effect of sharing the encoder for ADC and AOS tasks. We train these two tasks separately, i.e., setting ($\alpha$, $\beta$) in Eq.~\ref{eq:One-ASQP_loss} to (1.0, 0.0) and (0.0, 1.0).}  The results in Table~\ref{tab:ablation} show that our One-ASQP absorbs deeper information between two tasks and attains better performance.  By sharing the encoder and conducting joint training, the connection between the category and other sentiment elements can become more tightly integrated, thereby contributing to each other.

\begin{table*}[htp]
\footnotesize
\centering
\begin{tabular}{lcccccccc}
\hline 
\multirow{2}{*}{} & \multicolumn{2}{c}{\bf Restaurant-ACOS} & \multicolumn{2}{c}{\bf Laptop-ACOS} & \multicolumn{2}{c}{\bf en-Phone} & \multicolumn{2}{c}{\bf zh-FoodBeverage} \\ \cline{2-9}
 & ADC F1 & AOS F1 & ADC F1  & AOS F1 & ADC F1  & AOS F1 & ADC F1  & AOS F1 \\ \hline 
One-ASQP ($\alpha=1.0$, $\beta=0.0$)  & 68.64 & - & 47.45 & - & 63.43 & - & 64.57 & - \\
One-ASQP ($\alpha=0.0$, $\beta=1.0$) & - & 63.14 & - & 63.03 & - & 54.06 & - & 56.81 \\
One-ASQP (base) & \textbf{75.85} & \textbf{65.88} & \textbf{51.62} & \textbf{65.13} & \textbf{66.09} & \textbf{57.99} & \textbf{66.90} & \textbf{62.62} \\ \hline 
\end{tabular}
\caption{Ablation study of One-ASQP on two losses.}
\label{tab:ablation}
\end{table*}

\subsection{Effect of Different Quadruple Densities}\label{sec:effect_density}

We conduct additional experiments to test the effect of different quadruple densities.  Specifically, we keep those samples with only one quadruple in en-Phone and zh-FoodBeverage and construct two lower-density datasets, en-Phone (one) and zh-FoodBeverage (one).  We then obtain 1,528 and 3,834 samples in these two datasets, respectively, which are around one-fifth and two-fifth of the original datasets accordingly.  

{
We only report the results of our OneASQP with the base versions of the corresponding LMs and Paraphrase.  Results in Table~\ref{tab:quad_density_effect} show some notable observations: (1) Paraphrase can attain better performance on en-Phone (one) than our OneASQP.  It seems that generation-based methods are powerful in the low-resource scenario.  However, the performance is decayed in the full datasets due to the generation order issue.   (2) Our One-ASQP significantly outperforms Paraphrase in zh-FoodBeverage for both cases.  The results show that our OneASQP needs sufficient training samples to perform well.  However, in zh-FoodBeverage (one), the number of labeled quadruples is 3,834.  The effort is light in real-world applications.  

\begin{table}[htp]
\footnotesize
\centering
\begin{tabular}{l|cc|cc}
\hline 
\multirow{2}{*}{\textbf{Method}} & \multicolumn{2}{c|}{\textbf{en-Phone}} & \multicolumn{2}{c}{\textbf{zh-FoodBeverage}} \\ \cline{2-5} 
 & {{one}} & {{full}} & {{one}} & {{full}} \\ \hline 
Paraphrase & \textbf{49.78} & 48.23 & 49.23 & {50.23} \\
OneASQP & 36.12 & \textbf{53.58} & \textbf{53.39} & \textbf{57.79} \\ \hline 
\end{tabular}
\caption{Comparison results on different datasets with different quadruple densities.}
\label{tab:quad_density_effect}
\end{table}

\subsection{Error Analysis and Case Study}
To better understand the characteristics of our One-ASQP, especially when it fails.  We conduct the error analysis and case study in this section.  We check the incorrect quad predictions on all datasets and show one typical error example for each type from Laptop-ACOS in Fig.~\ref{fig:error_analysis}, where we report the percentage of errors for better illustration.  The results show that (1) In general, extracting aspects and opinions tends to introduce larger errors than classifying categories and sentiments.  Aspects and opinions have more complex semantic definitions than categories and sentiments, and extracting implicit cases further increases the difficulty of these tasks.  (2) There is a significant category error in Laptop-ACOS, likely due to an imbalance issue in which there are 121 categories with relatively small samples per category. For example, 35 categories have less than two quadruples.  (3) The percentage of opinion errors is higher than that of aspect errors in all datasets because opinions vary more than aspects, and there are implicit opinions in the new datasets.  This is reflected in the numbers of opinion errors in en-Phone and zh-FoodBeverage, which are 125 (37.31\%) and 395 (49.94\%), respectively, exceeding the corresponding aspect errors of 99 (29.55\%) and 246 (31.10\%). Removing samples with implicit opinions reduces the opinion errors to 102 and 260 in en-Phone and zh-FoodBeverage, indicating that explicit opinion errors are slightly larger than explicit aspect errors.  (4) The percentage of sentiment errors is relatively small, demonstrating the effectiveness of our proposed sentiment-specific horns tagging schema.

\begin{figure*}[htp]
    \begin{tabular}{cc}
\begin{minipage} {6cm}
\includegraphics[scale=0.05]{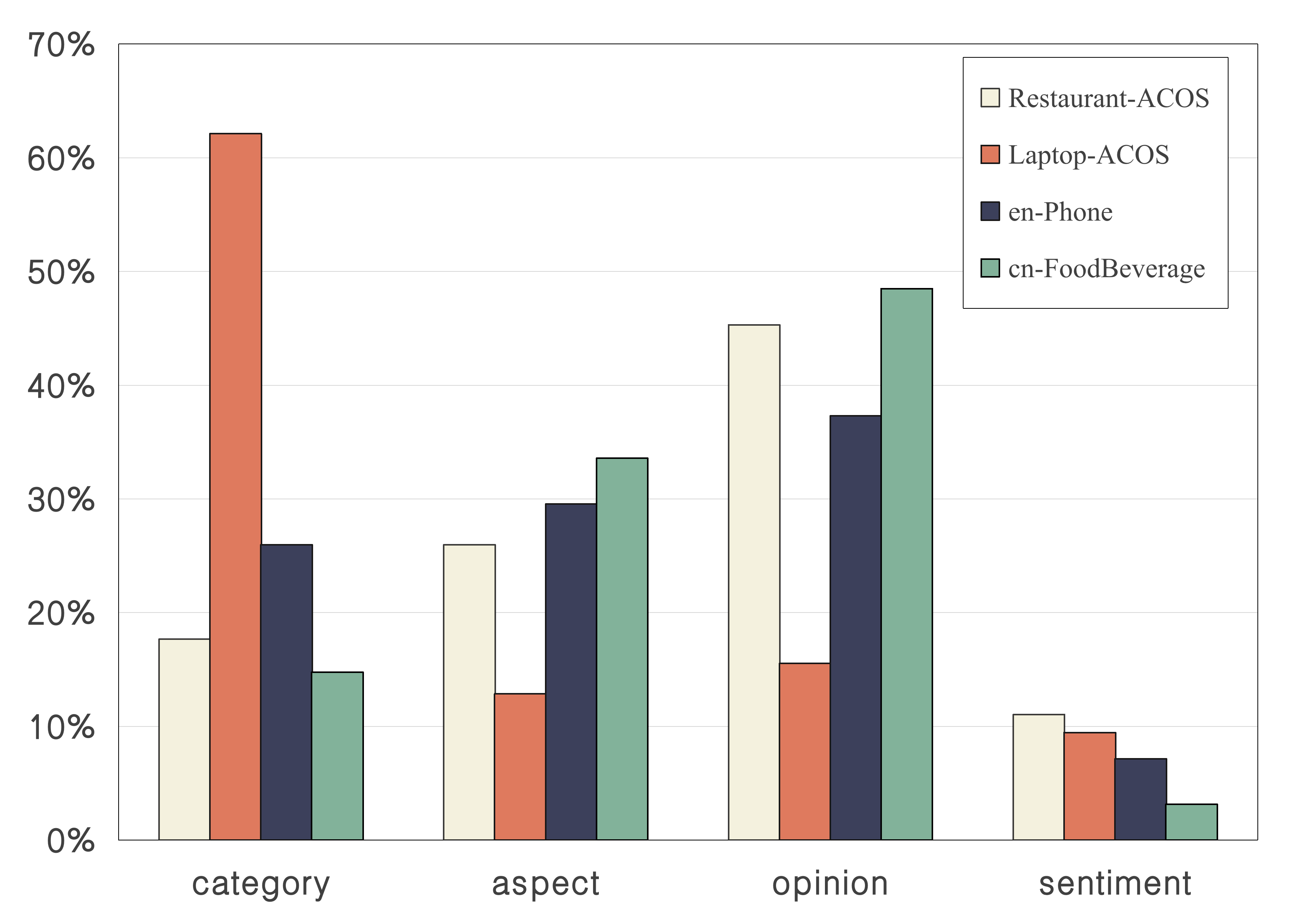}
\end{minipage}    
        & \begin{minipage} {8cm}
         \scriptsize
\begin{tabular}{|@{~}l@{~}|l@{~}|}
\hline
\textbf{Type} & \textbf{Example} \\\hline
\multirow{3}{*}{Category} & Input: the screen looked great.	\\
& Gold: (\underline{\blue{DISPLAY\#GENERAL}}, screen, great, POS) \\
& Pred.: (\underline{\blue{DISPLAY\#DESIGN\_FEATURES}}, screen, great, POS) \\\hline
\multirow{3}{*}{Aspect} 	& Input: works flawlessly and decent battery life.		\\
& Gold: (BATTERY\#OPERATION\_PERFORMANCE, \underline{\blue{battery}}, decent, POS) \\
& Pred.: (BATTERY\#OPERATION\_PERFORMANCE, \underline{\blue{battery life}}, decent, POS) \\\hline
\multirow{3}{*}{Opinion} & Input: the keyboard is backlit and big enough for my fingers. \\
& Gold: (KEYBOARD\#DESIGN\_FEATURES, keyboard, \underline{\blue{NULL}}, POS) \\
& Pred.: (KEYBOARD\#DESIGN\_FEATURES, keyboard, \underline{\blue{big}}, POS) \\\hline
\multirow{3}{*}{Sentiment} & Input: it starts up, runs without issues.	\\
& Gold: (OS\#OPERATION\_PERFORMANCE, starts up, NULL, \underline{\blue{NEU}}) \\
& Pred.: (OS\#OPERATION\_PERFORMANCE, starts up, NULL, \underline{\blue{POS}}) \\\hline 
\end{tabular}         
         \end{minipage}\\
  (a) Percentage of errors    & (b) Typical error examples from Laptop-ACOS.
    \end{tabular}
\caption{Error analysis and case study.  Though the predicted aspect and opinion differ from the golden ones in the above examples, they seem correct. } 
\label{fig:error_analysis} 
\end{figure*}

\section{Related Work}

{
\noindent\textbf{ABSA Benchmark Datasets} are mainly provided by the SemEval'14-16 shared challenges~\citep{DBLP:conf/semeval/PontikiGPPAM14,DBLP:conf/semeval/PontikiGPMA15,DBLP:conf/semeval/PontikiGPAMAAZQ16}.  The initial task is only to identify opinions expressed about specific entities and their aspects.  In order to investigate more tasks, such as AOPE, E2E-ABSA, ASTE, TASD, and ASQP, researchers have re-annotated the datasets and constructed some new ones~\citep{DBLP:conf/naacl/FanWDHC19,DBLP:conf/aaai/LiBLL19,DBLP:conf/emnlp/XuLLB20,DBLP:conf/aaai/WanYDLQP20,DBLP:conf/acl/CaiXY20}.  However, re-annotated datasets still contain the following limitations: (1) The data is collected from only one source, limiting the scope of the data; (2) the data size is usually small, where the maximum one is only around 4,000; (3) there is only a labeled quadruple per sentence and many samples share a common aspect, which makes the task easier; (4) the available public datasets are all in English.  The shortcomings of existing benchmark datasets motivate us to crawl and curate more data from more domains, covering more languages and with higher quadruple density.

{
\noindent\textbf{ASQP} aims to predict the four sentiment elements to provide a complete aspect-level sentiment structure~\citep{DBLP:conf/acl/CaiXY20,DBLP:conf/emnlp/ZhangD0YBL21}.  The task is extended to several variants, e.g., capturing the quadruple of holder-target-expression-polarity~\citep{DBLP:journals/corr/abs-2205-00440,DBLP:conf/semeval/LuR0L22} or the quadruple of target-aspect-opinion-sentiment in a dialogue~\citep{DBLP:journals/corr/abs-2211-05705}.  Existing studies can be divided into the pipeline or generation paradigm.  A typical {\em pipeline}-based work~\citep{DBLP:conf/acl/CaiXY20} has investigated different techniques to solve the subtasks accordingly.  It consits of (1) first exploiting double propagation (DP)~\citep{DBLP:journals/coling/QiuLBC11} or JET~\citep{DBLP:conf/emnlp/XuLLB20} to extract the aspect-opinion-sentiment triplets and after that, detecting the aspect category to output the final quadruples; (2) first utilizing TAS-BERT~\cite{DBLP:conf/aaai/WanYDLQP20} and the Extract-Classify scheme~\cite{DBLP:conf/aaai/WangPDX17} to perform the aspect-opinion co-extraction and predicting category-sentiment afterward.  Most studies fall in the {\em generation paradigm}~\citep{DBLP:conf/emnlp/ZhangD0YBL21,DBLP:conf/acl/MaoSYZC22,DBLP:conf/ijcai/BaoWJXL22,DBLP:conf/coling/GaoFLLLLBY22}.  ~\citet{DBLP:conf/emnlp/ZhangD0YBL21} is the first generation-based method to predict the sentiment quads in an end-to-end manner via a {\em PARAPHRASE} modeling paradigm.  It has been extended and overcome by Seq2Path~\citep{DBLP:conf/acl/MaoSYZC22} or tree-structure generation~\citep{DBLP:conf/acl/MaoSYZC22,DBLP:conf/ijcai/BaoWJXL22} to tackle the generation order issue or capture more information.  Prompt-based generative methods are proposed to assemble multiple tasks as LEGO bricks to attain task transfer~\citep{DBLP:conf/coling/GaoFLLLLBY22} or tackle few-shot learning~\citep{DBLP:journals/corr/abs-2210-06629}.   GEN-SCL-NAT~\citep{DBLP:journals/corr/abs-2211-07743} is introduced to exploit supervised contrastive learning and a new structured generation format to improve the naturalness of the output sequences for ASQP.  However, existing methods either yield error propagation in the pipeline-based methods or slow computation in the generation-based methods.  The shortcomings of existing methods motivate us to propose One-ASQP. 

\section{Conclusions}

In this paper, we release two new datasets, with the first dataset being in Chinese, for ASQP and propose One-ASQP, a method for predicting sentiment quadruples simultaneously.  One-ASQP utilizes a token-pair-based 2D matrix with sentiment-specific horns tagging, which allows for deeper interactions between sentiment elements, enabling efficient decoding of all aspect-opinion-sentiment triplets.  An elaborately designed ``[NULL]" token is used to identify implicit aspects or opinions effectively.  Extensive experiments are conducted to demonstrate the effectiveness and efficiency of One-ASQP.  Notably, existing strong baselines exhibit a decay in performance on the newly released datasets.  We hope these datasets and One-ASQP will inspire further technical development in this area.


\section*{Acknowledgments}
The work was partially supported by the IDEA Information and Super Computing Centre (ISCC) and the National Nature Science Foundation of China (No. 62201576).

\section*{Limitations}
Our proposed One-ASQP still contains some limitations:
\begin{compactitem}
\item Our One-ASQP does not solve the case of IA\&IO.  We defer the technical exploration of this issue to future work.
\item One-ASQP has to split the ASQP task into two subtasks, ADC and AOSC.  It is still promising to explore more effective solutions, e.g., by only one task, which can absorb deeper interactions between all elements.
\item Generally, One-ASQP suffers more opinion errors than other sentiment elements due to the fine-grained annotation and implicit opinions issues.  It is possible to tackle it by exploring more advanced techniques, e.g., syntax or semantics augmentation, to dig out deeper connections between options and other sentiment elements.
\item One-ASQP tends to make errors when there are many aspect categories with small labeled quadruples.  It is also significant to explore more robust solutions to detect the aspect categories in the low-resource  scenario. 
\item Though we have released datasets in both English and Chinese, we do not explore ASQP in the multi-lingual scenario.  We leave this as future work. 
\end{compactitem}

\section*{Ethics Statement}
We follow the ACL Code of Ethics.  In our work, there are no human subjects and informed consent is not applicable.  

\bibliography{anthology,custom}
\bibliographystyle{acl_natbib}

\appendix
\section{More Details about Datasets Construction}\label{app:datasets}
This section provides more details about constructing the two datasets, en-Phone and zh-FoodBeverage. 

\begin{table*}[htp]
\small
\centering
    \begin{tabular}{l@{~~}|l@{~~}c}
    \hline
    &\multicolumn{1}{c}{Sentence} & Labeled Quadruples \\\hline
   Ex.~1 & \begin{minipage}{6.4cm}
       \begin{tabular}{@{}p{6.4cm}}
        This \textcolor{aspect}{screen} is \textcolor{opinion}{good overall}, although the \textcolor{aspect}{screen size} is \textcolor{opinion}{not large}, but looks \textcolor{opinion}{very} \textcolor{aspect}{clear}
       \end{tabular}
   \end{minipage}
& \begin{minipage}{8cm}
\begin{tabular}{l}
 (\textcolor{category}{Screen\#General}, \textcolor{aspect}{screen}, \textcolor{opinion}{good overall}, \textcolor{sentiment}{POS}) \\
(\textcolor{category}{Screen\#Size}, \textcolor{aspect}{screen size}, \textcolor{opinion}{not large}, \textcolor{sentiment}{NEG})\\
(\textcolor{category}{Screen\#Clarity}, \textcolor{aspect}{clear}, \textcolor{opinion}{very}, \textcolor{sentiment}{POS}) 
\end{tabular}
\end{minipage}
\\\hline
    Ex.~2 & \textcolor{opinion}{Don’t like} \textcolor{aspect}{face recognition} and \textcolor{aspect}{battery life}.
    & 
    \begin{minipage}{8cm}
        \begin{tabular}{p{8cm}}
            (\textcolor{category}{Security\#Screen Unlock}, \textcolor{aspect}{face recognition}, \textcolor{opinion}{Don’t like}, \textcolor{sentiment}{NEG})\\
            (\textcolor{category}{Battery/Longevity\#Battery life}, \textcolor{aspect}{battery life}, \textcolor{opinion}{Don't like}, \textcolor{sentiment}{NEG})
        \end{tabular}
    \end{minipage}\\\hline
   Ex.~3 &   \textcolor{opinion}{Very fast} \textcolor{aspect}{delivery} \& \textcolor{aspect}{phone} is \textcolor{opinion}{working well}.        
    & 
      \begin{minipage}{8.cm}
        \begin{tabular}{p{8.cm}}
        (\textcolor{category}{Logistics\#Speed}, \textcolor{aspect}{delivery}, \textcolor{opinion}{Very fast}, \textcolor{sentiment}{POS})\\
        (\textcolor{category}{Overall Rating\#General}, \textcolor{aspect}{phone}, \textcolor{opinion}{working well}, \textcolor{sentiment}{POS})
      \end{tabular}
    \end{minipage}
      \\\hline
    Ex.~4 &  \begin{minipage}{6.4cm}
        \begin{tabular}{@{}p{6.4cm}}
       It's \textcolor{opinion}{very} \textcolor{aspect}{cheap}.  The first time I bought the phone I wanted. 
       \end{tabular}
    \end{minipage}
    & 
      \begin{minipage}{8.cm}
        \begin{tabular}{p{8.cm}}
       (\textcolor{category}{Price\#General}, \textcolor{aspect}{cheap}, \textcolor{opinion}{very}, \textcolor{sentiment}{POS})
       \end{tabular}
    \end{minipage}
       \\\hline       
    \end{tabular}
    \caption{Three opinionated sentences and the labeled quadruples.  }
    \label{tab:ex_quadruples}
\end{table*}

\subsection{Data Sources}\label{app:source}
The English ASQP dataset, en-Phone, is collected from reviews on Amazon UK~\footnote{\url{https://www.amazon.co.uk/}}, Amazon India~\footnote{\url{https://www.amazon.in/}} and Shopee~\footnote{\url{https://shopee.com.my/}} in July and August of 2021, covering 12 cell phone brands, such as Samsung, Apple, Huawei, OPPO, Xiaomi, etc.  

The first Chinese ASQP dataset, zh-FoodBeverage, is collected from the Chinese comments on  forums~\footnote{\url{http://foodmate.net/}}, Weibo~\footnote{\url{https://weibo.com/}}, news~\footnote{\url{https://chihe.sohu.com/}} and e-commerce platforms~\footnote{\url{https://www.jd.com/},\url{https://www.taobao.com/}} in the years 2019-2021 under the categories of Food and Beverage.

%


\subsection{Annotation Guidelines}\label{app:guideline}

The following outlines the guidelines for annotating the four fundamental sentiment elements of ASQP and their outcomes.  It can be noted that our labeled ASQP quadruples are more fine-grained and more difficult than those in existing ASQP benchmark datasets.

\subsubsection{Aspect Categories}\label{app:categories}
The \textbf{aspect category} defines the type of the concerned aspect.  Here, we apply a two-level category system, which is defined by our business experts for the sake of commercial value and more detailed information.  For example, ``Screen'' is a first-level category.  It can include second-level categories, such as ``Clarity'', ``General'', and ``Size'', to form the final second-level categories as ``Screen\#Clarity'', ``Screen\#General'', and ``Screen\#Size''.  In the experiments, we only consider the second-level categories.

As reported in Table~\ref{tab:datasets}, the number of categories for en-Phone and zh-FoodBeverage is 88 and 138, respectively.  The number of labeled quadruples per category is larger than 5.  Though Laptop-ACOS consists of 121 categories, if we filter out the categories with less than 5 annotated quadruples, the number of categories is reduced to 75.  Hence, we provide more dense and rich datasets for ASQP. 

\subsubsection{Aspect Terms}\label{app:aspect}
The \textbf{aspect} term is usually a noun or a noun phrase, indicating the opinion target, in the text.  It can be implicit in a quadruple~\citep{DBLP:conf/acl/CaiXY20}.  For the sake of commercial analysis, we exclude sentences without aspects.  Moreover, to provide more fine-grained information, we include three additional rules: 
\begin{compactitem}
    \item The aspect term can be an adjective or verb when it can reveal the sentiment categories.  For example, as the example of en-Phone in Table~\ref{tab:ex_illustration}, ``recommended'' is also labeled as an aspect in ``Highly recommended'' because it can identify the category of ``Buyer\_Atitude\#Willingness\_Recommend''.  In Ex.~1 and Ex.~4 of Table~\ref{tab:ex_quadruples}, ``clear'' and ``cheap'' are labeled as the corresponding aspect terms because they can specify the category of ``Screen\#Clarity'' and ``Price\#General'', accordingly.
    \item Pronoun is not allowed to be an aspect term as it cannot be identified by the quadruples only.  For example, in the example of ``pretttyyyy and affordable too!!! I love it!!  Thankyouuu!!'', ``it'' cannot be labeled as the aspect though we know it indicates a phone from the context.  
    \item Top-priority in labeling fine-grained aspects.  For example, in the example of ``Don't purchase this product'', ``purchase'' is more related to a customer's purchasing willingness while ``product'' is more related to the overall comment, we will label ``purchase'' as the aspect.  
\end{compactitem}

\begin{CJK*}{UTF8}{gbsn}
\begin{table*}[htpb]
\scriptsize
\centering
\begin{tabular}
{@{~}l@{~}c@{~~}c
}
\hline
Source & Opinionated sentences & Quadruples \\\hline
{en-Phone} 
&  
\begin{minipage}{5.4cm}
\begin{tabular}{p{5.4cm}}
{\textcolor{aspect}{Item received} \textcolor{opinion}{took a long time}. \textcolor{aspect}{Looks} \textcolor{opinion}{nice} and \textcolor{opinion}{good} \textcolor{aspect}{quality} too. \textcolor{aspect}{Price} is \textcolor{opinion}{cheaper than retail costs}. Bought it for my \textcolor{aspect}{mom} and she \textcolor{opinion}{likes} it! \textcolor{opinion}{Highly} \textcolor{aspect}{recommended}.} \\\\
\end{tabular}
 \end{minipage}
 & 
\begin{minipage}{7.cm}
\begin{tabular}{p{7.cm}}
    (\textcolor{category}{Logistics\#Logistics speed}, \textcolor{aspect}{Item received}, \textcolor{opinion}{took a long time}, \textcolor{sentiment}{NEG}), \\
 (\textcolor{category}{Exterior Design\#Aesthetics}, \textcolor{aspect}{Looks}, \textcolor{opinion}{nice}, \textcolor{sentiment}{POS}), \\
  (\textcolor{category}{Product Quality\#General}, \textcolor{aspect}{quality}, \textcolor{opinion}{good}, \textcolor{sentiment}{POS}), \\
(\textcolor{category}{Price\#General}, \textcolor{aspect}{Price}, \textcolor{opinion}{cheaper than retail costs}, \textcolor{sentiment}{POS}), \\
  (\textcolor{category}{Audience\#Users}, \textcolor{aspect}{mom}, \textcolor{opinion}{likes}, \textcolor{sentiment}{POS}), \\
 (\textcolor{category}{Buyer attitude\#Recommendable}, \textcolor{aspect}{recommended}, \textcolor{opinion}{Highly}, \textcolor{sentiment}{POS})
 \end{tabular}
 \end{minipage}
\\ \hline
{\multirow{1}{*}{zh-FoodBeverage}} & 
\begin{minipage}{5.4cm}
\begin{tabular}{p{5.4cm}}
我挑选这个品牌的主要原因是它这里\textcolor{opinion}{含}一项\textcolor{aspect}{乳铁蛋白}\textcolor{opinion}{促进}\textcolor{aspect}{宝宝}吸收的，所以宝宝喝它\textcolor{opinion}{没有}\textcolor{aspect}{奶瓣}的情况，同时它的\textcolor{aspect}{口味}也\textcolor{opinion}{接近母乳}，\textcolor{aspect}{宝宝}\textcolor{opinion}{很爱喝}。可惜的是，买了那么多奶粉，\textcolor{aspect}{赠品}\textcolor{opinion}{没收到}，也不知道哪个环节出的问题，连\textcolor{aspect}{赠品}的影子都\textcolor{opinion}{没看见}，只能认倒霉了，无处对证去
\\  The main reason I picked this brand is that it \textcolor{opinion}{contains} \textcolor{aspect}{Lactoferrin} to \textcolor{opinion}{promote} \textcolor{aspect}{baby's absorption}, so the baby drinks it \textcolor{opinion}{without} a \textcolor{aspect}{milk valve}, while its \textcolor{aspect}{taste} is also \textcolor{opinion}{close to breast milk}, the \textcolor{aspect}{baby} \textcolor{opinion}{loves to drink}.  Unfortunately, I bought so much milk powder and \textcolor{opinion}{did not receive} a \textcolor{aspect}{gift}, and I do not know which part of the problem, even the shadow of the \textcolor{aspect}{gift} \textcolor{opinion}{did not see}, can only admit bad luck, there is no evidence to testify. 
\end{tabular}
 \end{minipage} 
 & 
 \begin{minipage}{7.cm}
\begin{tabular}{p{7.cm}}
    (\textcolor{category}{成分\#营养成分}, \textcolor{aspect}{乳铁蛋白}, \textcolor{opinion}{含}, \textcolor{sentiment}{POS}), \\     
    (\textcolor{category}{营养\#吸收}, \textcolor{aspect}{宝宝吸收}, \textcolor{opinion}{促进}, \textcolor{sentiment}{POS}), \\
    (\textcolor{category}{不良反应\#其他不适}, \textcolor{aspect}{奶瓣}, \textcolor{opinion}{没有}, \textcolor{sentiment}{POS}), \\
    (\textcolor{category}{味道\#综合味道}, \textcolor{aspect}{口味}, \textcolor{opinion}{接近母乳}, \textcolor{sentiment}{POS}), \\
    (\textcolor{category}{使用\#受众群体}, \textcolor{aspect}{宝宝}, \textcolor{opinion}{很爱喝}, \textcolor{sentiment}{POS}), \\
    (\textcolor{category}{促销\#赠品}, \textcolor{aspect}{赠品}, \textcolor{opinion}{没收到}, \textcolor{sentiment}{NEG}), \\
    (\textcolor{category}{促销\#赠品}, \textcolor{aspect}{赠品}, \textcolor{opinion}{没看见}, \textcolor{sentiment}{NEG}), \\
     (\textcolor{category}{Ingredients\#Nutritional Composition}, \textcolor{aspect}{lactoferrin}, \textcolor{opinion}{contains}, \textcolor{sentiment}{POS}), \\
     (\textcolor{category}{Nutrition\#Absorption}, \textcolor{aspect}{baby's absorption}, \textcolor{opinion}{promote}, \textcolor{sentiment}{POS}), \\
     (\textcolor{category}{Adverse reactions\#Other discomfort}, \textcolor{aspect}{milk valve}, \textcolor{opinion}{without}, \textcolor{sentiment}{POS}), \\
     (\textcolor{category}{Flavor\#Comprehensive flavor}, \textcolor{aspect}{taste}, \textcolor{opinion}{close to breast milk}, \textcolor{sentiment}{POS}), \\
     (\textcolor{category}{Use\#Audience group}, \textcolor{aspect}{baby}, \textcolor{opinion}{loves to drink}, \textcolor{sentiment}{POS}), \\
     (\textcolor{category}{Promotion\#Giveaway}, \textcolor{aspect}{gift}, \textcolor{opinion}{did not receive}, \textcolor{sentiment}{NEG}), \\
     (\textcolor{category}{Promotion\#Giveaway}, \textcolor{aspect}{gift}, \textcolor{opinion}{did not see}, \textcolor{sentiment}{NEG}), \\

 \end{tabular}
 \end{minipage}
\\ \hline
\end{tabular}
\caption{Typical examples of the labeled quadruples in en-Phone and zh-FoodBeverage.
\label{tab:ex_illustration}}
\end{table*}
\end{CJK*}
\subsubsection{Opinion Terms}\label{app:opinion}
The \textbf{opinion} term describes the sentiment towards the aspect.  {An opinion term is usually an adjective or a phrase with sentiment polarity. }  Here, we include more labeling criteria:
\begin{compactitem}
    \item When there is a negative word, e.g., ``Don't'', ``NO'', ``cannot'', ``doesn't'', the negative word should be included in the opinion term.  For example, ``not large'' and ``Don’t like'' are labeled as the corresponding opinion terms in Ex.~1 and Ex.~2 of Table~\ref{tab:ex_quadruples}
    \item When there is an adverb or a preposition, e.g., ``very'', ``too'', ``so'', ``inside'', ``under'', ``outside'', the corresponding adverb or preposition should be included in the opinion term.  For example, in Ex.~3 of Table~\ref{tab:ex_quadruples}, ``Very fast'' is labeled as an  opinion term.  Usually, in Restaurant-ACOS and Laptop-ACOS, ``Very'' is not included in the opinion term.  Moreover, in Ex.~1 of Table~\ref{tab:ex_quadruples}, ``very'' in ``very clear'' is labeled as an opinion term while in Ex.~4, ``very'' in ``very cheap'' is labeled as the opinion term.  
\end{compactitem}
These examples show that our labeled opinion terms are more fine-grained and complicated, but more valuable for real-world applications.  This increases the difficulty of extracting opinion terms and demonstrates the significance of our released datasets to the ASBA community.

\subsubsection{Sentiment Polarity}\label{app:SP}
The \textbf{sentiment polarity} belongs to one of the sentiment classes, \{POS, NEU, NEG\}, for the positive, neutral, and negative sentiment, respectively.  In zh-FoodBeverage, for commercial considerations, we only label sentences with positive and negative sentiments and exclude those with neutral sentiment.  

\subsection{Quadruple Density Analysis}\label{app:quads_density}

\begin{figure}[ht] 
\centering 
\includegraphics[scale=0.06]{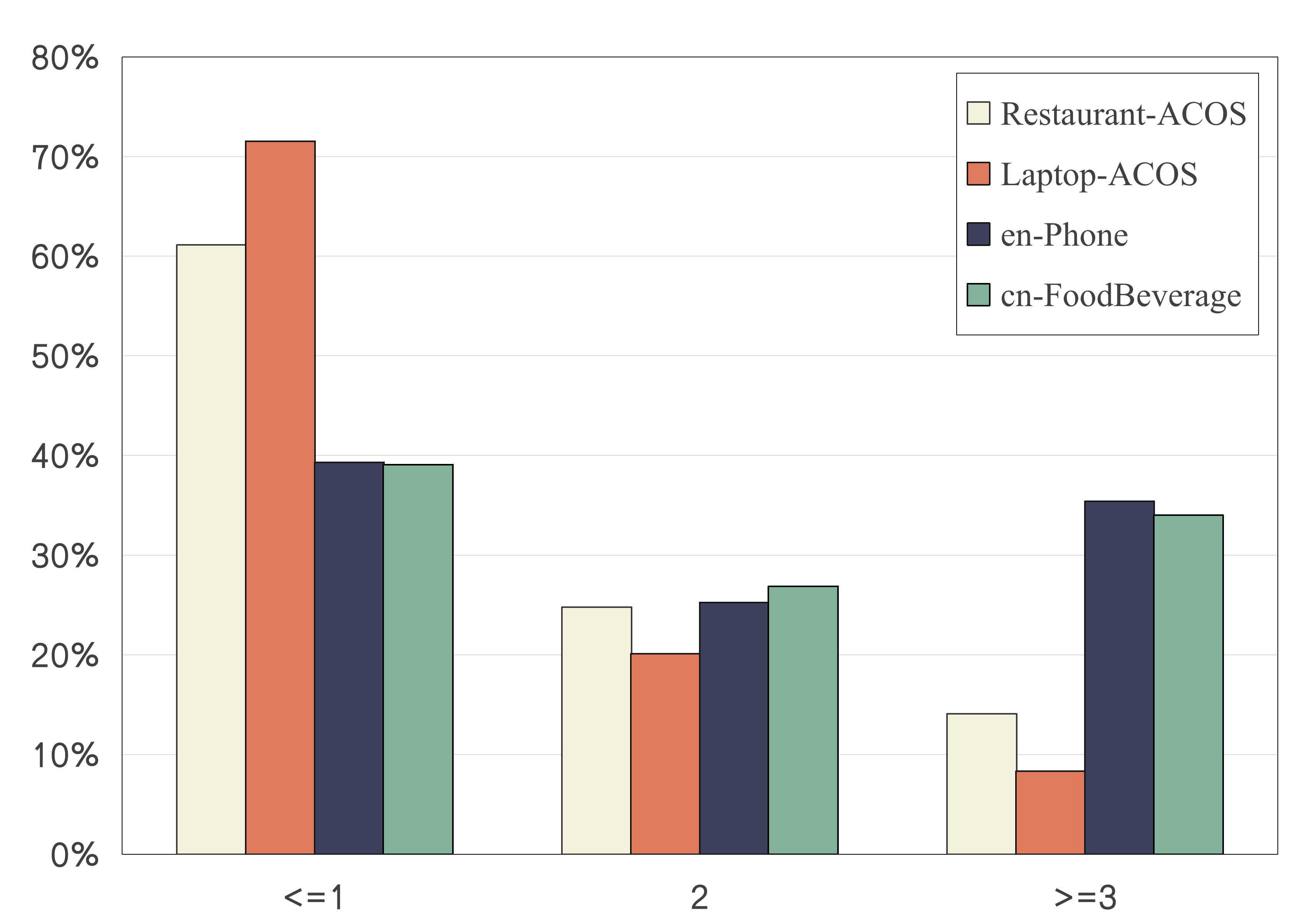}
\caption{The ratio of the number of quadruples per sentence in four datasets. } 
\label{fig:quads_num} 
\end{figure}

\begin{table*}[htp]
\small
\centering
\begin{tabular}{l@{~}|c@{~}c@{~}|c@{~}c@{~}|c@{~}c@{~}|c@{~}c@{~}}
\hline
\multirow{2}{*}{\textbf{Method}} & \multicolumn{2}{c|}{\textbf{Restaurant-ACOS}} & \multicolumn{2}{c|}{\textbf{Laptop-ACOS}} & \multicolumn{2}{c|}{\textbf{en-Phone}} & \multicolumn{2}{c}{\textbf{zh-FoodBeverage}}\\ \cline{2-9} 
& {{Train}} & {{Inference}} & {{Train}} & {{Inference}} & {{Train}} & {{Inference}} & {{Train}} & {{Inference}} \\ \hline
Extract-Classify & 38.43 & 14.79 & 72.25 & 20.23 & 158.34 & 25.23 & 301.42 & 70.34 \\
Paraphrase & 30.52 & 58.23 & 59.23 & 69.23 & 99.23 & 160.56 & 664.23 & 673.32 \\ 
GEN-SCL-NAT & 35.32 & 61.64 & 63.53 & 72.23 & 104.23 & 175.23 & 748.56 & 706.43 \\ 
\hline
OneASQP (base) & 11.23 & 6.34 (29.35) & 19.03 & 8.34 (39.83) & 32.23 & 6.32 (35.45) & 71.23 & 13.23 (31.74) \\
OneASQP (large) & 17.63 & 14.63 (44.62) & 36.63 & 8.45 (49.45) & 105.23 & 10.34 (61.23) & 140.23 & 30.46 (56.32) \\ 
\hline
\end{tabular}
\caption{Time cost in seconds on all datasets.  For a fair comparison with baselines, we record our One-ASQP  inference time when setting the batch size to 1 and report them in the round bracket.  The default batch size is 32. }
\label{tab:all_time}
\end{table*}
For better illustration, we count the number of quadruples per sentence in four datasets and show the ratios in Fig.~\ref{fig:quads_num}.  It is shown that (1) In terms of sentences with at most one labeled quadruple, Restaurant-ACOS contains 61.12\% of the sentences and it is 71.54\% in Laptop-ACOS.  Meanwhile, it is 39.33\% and 39.10\% in en-Phone and zh-FoodBeverage, respectively.  (2) In terms of sentences with at least three labeled quadruples, it drops significantly to 14.09\% in Restaurant-ACOS and 8.34\% in Laptop-ACOS.  Meanwhile, it is 35.19\% in en-Phone and 34.01\% in zh-FoodBeverage.  Hence, our released datasets are more dense and balanced.

\section{More Experimental Results}\label{app:more_exp}

\subsection{Computation Efficiency}\label{app:efficiency}

Table~\ref{tab:all_time} reports the time cost (in seconds) on all four datasets.  The base versions of the corresponding LMs are applied in  Extract-Classify.  It shows that One-ASQP is efficient in both training and inference, which is a favorite for real-world deployment. 

\begin{table}[htp]
\small
\centering
\begin{tabular}{llclclcl}
\hline
\multicolumn{2}{c}{} & \multicolumn{2}{c}{Variant~1} & \multicolumn{2}{c}{Variant~2} & \multicolumn{2}{c}{One-ASQP} \\ \hline
\multicolumn{2}{l}{\textbf{Restaurant-ACOS}} & \multicolumn{2}{c}{58.39} & \multicolumn{2}{c}{57.23} & \multicolumn{2}{c}{\textbf{59.78}} \\
\multicolumn{2}{l}{\textbf{Laptop-ACOS}} & \multicolumn{2}{c}{41.05} & \multicolumn{2}{c}{39.12} & \multicolumn{2}{c}{\textbf{41.37}} \\
\multicolumn{2}{l}{\textbf{en-Phone}} & \multicolumn{2}{c}{51.23} & \multicolumn{2}{c}{49.72} & \multicolumn{2}{c}{\textbf{53.58 }} \\ 
\multicolumn{2}{l}{\textbf{zh-FoodBeverage}} & \multicolumn{2}{c}{57.23} & \multicolumn{2}{c}{55.95} & \multicolumn{2}{c}{\textbf{57.79}} \\ \hline 
\end{tabular}
\caption{Comparison of One-ASQP with two other variants for ASQP.}
\label{exp:interaction_result}
\end{table}

\subsection{Effect of Variants of Interactions}
Though our One-ASQP separates the task into ACD and AOSC.  There are still other variants to resolve the ASQP task.  Here, we consider two variants:

{\bf Variant~1}: The ASQP task is separated into three sub-tasks: aspect category detection (ACD), aspect-opinion pair extraction (AOPC), and sentiment detection.  More specifically, ACD and sentiment detection are fulfilled by classification models.  For AOPC, we adopt the sentiment-specific horns tagging schema proposed in Sec.~\ref{sec:AOSC}.  That is, we only co-extract the aspect-opinion pairs.  In the implementation, we set the tags of AB-OE-*SENTIMENT to AB-EO and reduce the number of tags for AOSC to three, i.e., \{AB-OB, AE-OE, AB-OE\}.  

{\bf Variant~2}: We solve the ASQP task by a unified framework.  Similarly, via the sentiment-specific horns tagging schema proposed in Sec.~\ref{sec:AOSC}, we extend the tags of AB-EO-*SENTIMENT to AB-OE-*SENTIMENT-*CATEGORY.  Hence, the number of tags increases from 5 to $2+|\gS|*|\gC|$, where $|\gS|$ is the number of sentiment polarities and $|\gC|$ is the number of categories.  This setting allows us to extract the aspect-opinion pairs via the 2D matrix while decoding the categories and sentiment polarities via the tags. 


Tabel~\ref{exp:interaction_result} reports the compared results on four datasets, where the base versions of the corresponding LMs are applied.  The results show that (1) our One-ASQP performs the best over the proposed two variants.  We conjecture that the aspect-opinion-sentiment triplets are in a suitable tag space and our One-ASQP can absorb their interactions effectively.  (2) Variant~2 performs the worst among all results.  We conjecture that the search tag space is too large and the available datasets do not contain enough information to train the models. 


\end{document}